\documentclass{article}

\usepackage{verbatim}
\usepackage{amsmath}
\usepackage{amssymb}
\usepackage{natbib}
\newcommand{\tends}{\rightarrow}
\newcommand\argmin{\mathop{\mbox{{\rm argmin}}}\limits}

\newcommand\E{\mathop{\mbox{{\rm E}}}\limits}
\newtheorem{example} {Example}

\usepackage{graphicx}
\usepackage{subfig}
\usepackage{algorithm}
\usepackage{algorithmic}

\bibpunct{(}{)}{;}{a}{,}{,}

\author{
Yi-Hao Kao\footnote{Corresponding author contact: yhkao@alumni.stanford.edu}  \hspace{0.2in} Benjamin Van Roy \\
\\
\emph{Stanford University}
}
\title{Directed Time Series Regression for Control}
\date{January 30, 2011}

\begin{document}

\maketitle

\begin{abstract} 
We propose {\it directed time series regression}, a new approach to estimating parameters of time-series models for use in certainty equivalent model predictive control. The approach combines merits of least squares regression and empirical optimization.  Through a computational study involving a stochastic version of a well known inverted pendulum balancing problem, we demonstrate that directed time series regression can generate significant improvements in controller
performance over either of the aforementioned alternatives.
\end{abstract} 

\section{Introduction}
A common approach to stochastic control, sometimes referred to as {\it certainty equivalent model predictive control},  involves at each time forecasting future outcomes of exogenous random variables and optimizing control actions over a planning horizon under the assumption that these forecasts will be realized \citep{Bertsekas08, Box08}.  The first of the sequence of actions is taken. Forecasting and optimization are repeated at each subsequent time step.  In this paper, we develop a regression algorithm that fits time series forecasting models for use in such a context.

We focus attention on linear systems with quadratic cost, though the issues and methods we introduce generalize.  In particular, we will consider a dynamic system that evolves over discrete time steps $t = 0,1, 2,\ldots$.  At each time $t$, the state $x_t \in \mathbb{R}^P$ is updated according to
$$x_{t+1} = A x_t + B u_t + C w_t,$$
where $u_t \in \mathbb{R}^Q$ is a control action, $w_t \in \mathbb{R}^S$ is a random disturbance, and $A$, $B$, and $C$ are known matrices of appropriate dimension. Each control action $u_t$ is selected just before the disturbance $w_t$ is observed. The objective is to minimize average expected cost
$$ \lim_{h \tends \infty} \E\left[\frac{1}{h} \sum_{t=0}^{h-1} g( x_{t}, u_{t}) \right],$$
where the per period cost function is a positive definite quadratic of the form $g(x,u) = x^\top G_1 x + x^\top G_2 u + u^\top G_3 u$, for some $G_1$, $G_2$, and $G_3$. The following example illustrates a specific context.

\begin{figure}[ht]
\vskip 0.2in
\begin{center}
\centerline{\includegraphics[scale=0.4]{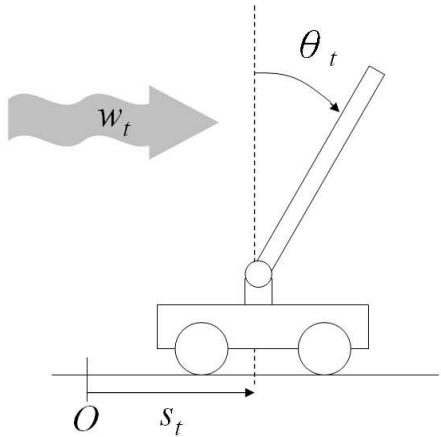}}
\caption{Inverted pendulum on a cart.}
\label{fig:invpend}
\end{center}
\vskip -0.2in
\end{figure}

\begin{example}
Consider a problem of balancing an inverted pendulum on a cart, as illustrated in Figure \ref{fig:invpend}.  Take the state to be a four-dimensional vector $x_t = [s_t \ \ \theta_t \ \ \dot{s}_t \ \ \dot{\theta}_t]^\top$, where $s_t$ is the position of the cart, $\theta_t$ is the angle of the pendulum, and $\dot{s}_t$ and $\dot{\theta}_t$ are their rates of change.  The cart can move freely along the horizontal axis, and is controlled through applying a voltage $u_t$ to its motor.  We discretize time, and consider a linearization of the system dynamics around the balance point as in \citet{Landry05}, which in the absence of disturbances can be written as $x_{t+1} = A x_t + B u_t$ for some $A$ and $B$. We introduce to the system a disturbance $w_t$ which can be thought of as a force exerted by a gust of wind at time $t$.  With the disturbances, the linearized system equation becomes $x_{t+1} = A x_t + B u_t + C w_t$ for some $C$. The objective is to center the cart and balance the pole while minimizing energy expenditure, and this is represented by a per period cost function $g(x_t,u_t) = x_t^\top G_1 x_t + G_3 u_t^2$, for some $G_1$ and $G_3$, where the first term captures how far the current state is from being centered and balanced and the second term reflects the energy applied by the control action.
\end{example}

This example may seem simple and specialized, but our framework captures a broad set of applications.  For example, we could model the control of a robot that receives instructions from a human user by encoding in $x_t$ both the physical state and a tentative plan.  Then, $w_t$ could represent changes to the plan communicated by the user and $u_t$ could represent an action taken by the robot as it tries to follow the plan.

An important issue is how future disturbances are forecasted.  In the inverted pendulum problem, for example, accurate predictions of future wind patterns facilitate more effective control.  We consider the use of linear time series models with coefficients estimated based on a historical sequence of disturbances.  The main issue we focus on in this paper is how one should estimate coefficients. Note that the setting of linear systems with quadratic cost and linear time series model is one of the most commonly used in applied work and also offers a simple starting point for exploring how learning and control should be coordinated.  

Given a linear model, it is common to estimate coefficients by minimizing squared error over historical data and then use the model with the resulting coefficients to generate forecasts for model predictive control. However, as we will discuss further, least squares regression leaves room for improvement when the observed process is not itself generated by the specified time series model.  In particular, in such situations it can be beneficial to take the control objective into account when computing coefficients.

One approach that does this is empirical optimization, sometimes referred to as empirical risk minimization, which computes coefficients that would have minimized historical average cost given the historical sequence of disturbances.  Under certain technical assumptions, if an infinite history of observations is available, empirical optimization yields coefficients that minimize future average expected cost.  However, with finite data, empirical optimization often works
poorly because it overspecializes to the data.

The main contribution of this paper is a new approach to fitting coefficients which we refer to as {\it directed time series regression}. This approach combines merits of least squares regression and empirical optimization. As we will demonstrate through a computational study of the inverted pendulum balancing problem, using directed time series regression can lead to large improvements in controller performance over either of the aforementioned alternatives.  In relation to prior work,
directed time series regression can be viewed as an extension of directed regression \citep{Kao09} to a context involving forecasting and control. 
This extension builds on several new contributions relative to what is presented in their work.  First, dynamic control problems are more complex than the static decision problem 
addressed in their work, since a state process drives decisions and can become unstable whereas there is no notion of evolving state in static problems.
Second, the empirical optimization problem considered in their work is a convex quadratic optimization problem, while the empirical optimization problem in our new context is not convex. We propose efficient approximation algorithms to address this problem. Third, the cross-validation technique employed in the algorithm of \citet{Kao09} does not apply in our new time series context, and we devise a sliding window cross-validation technique for this purpose.

\section{Certainty Equivalent Model Predictive Control}

Recall that the problem at hand involves controlling a system that evolves according to 
\begin{equation} \nonumber
x_{t+1} = A x_t + B u_t + C w_t,
\end{equation}
with a goal of minimizing average expected cost, where the per period cost function is a positive definite quadratic of the form $g(x,u) = x^\top G_1 x + x^\top G_2 u + u^\top G_3 u$.

Although the time horizon of interest is infinite, to simplify the problem, at each time $t$, model predictive control (MPC) aims to optimize a finite horizon objective of the form
$${\E}_t\left[ \sum_{\tau=t}^{\tau=t+M} g( x_{\tau}, u_{\tau} )\right],$$
where $M$ is the horizon time and the subscript $t$ indicates that the expectation is conditioned on $x_t, w_{t-1}, w_{t-2}, w_{t-3}, \ldots$.  After deriving a control policy that optimizes
this objective, MPC applies the policy at time $t$ to generate a control action $u_t$.  Then, after observing $w_t$ and $x_{t+1}$, a new finite horizon problem is formulated with a horizon spanning times $t+1$ through $t+1+M$.  This new problem is solved to produce a control action $u_{t+1}$.  The process is repeated at each subsequent time step.

In order to compute expected future costs, MPC requires a model that provides a distribution over future disturbance sequences.  Certainty equivalent MPC simplifies the problem by assuming that future disturbances follow a deterministic sequence $\hat{w}_{t,t}, \hat{w}_{t,t+1}, \hat{w}_{t,t+2}, \ldots$ generated at time $t$ by a forecasting model.  Hence, the objective of the optimization problem solved at time $t$ becomes
\begin{equation}
\sum_{\tau=t}^{\tau=t+M} g( x_{\tau}, u_{\tau} ),  \label{eq:mpcobj}
\end{equation}
where the state dynamics are governed by the same linear difference equation as before but with disturbances given by $(w_t, w_{t+1}, w_{t+2}, \ldots) = (\hat{w}_{t,t}, \hat{w}_{t,t+1}, \hat{w}_{t,t+2}, \ldots)$.

For our context, which involves a linear system and a quadratic cost function, certainty equivalent MPC generates a control policy that is linear in state and forecasts.  To simplify our discussion we will assume that disturbances are scalar-valued, though the ideas and methods we will present  
extend to the vector-valued case.  Under this assumption, there are matrices $L$ and $H$, which depend only on $A$, $B$, $C$, $G_1$, $G_2$, $G_3$, and $M$, such that at each time $t$, the action
employed by certainty equivalent MPC is given by
\begin{equation}
u_t = L x_t + H \hat{w}_{t,t}^{t,t+M-1}, \label{eq:mu}
\end{equation}
where $\hat{w}_{t,t}^{t,t+M-1} = [\hat{w}_{t,t} \ \cdots \ \hat{w}_{t,t+M-1}]^\top$. In particular, this action minimizes the objective in (\ref{eq:mpcobj}).

Certainty equivalent MPC leads to effective decisions for a wide variety of control problems.  For example, in our context of linear-quadratic control, if the process that generates disturbances is known and ergodic and each forecast $\hat{w}_{t,t+\tau}$ is equal to the expected disturbance 
${\E}_t[w_{t+\tau}]$ then the performance of certainty equivalent MPC becomes arbitrarily close to optimal as $M$ grows.  Certainty equivalent MPC has also been applied with success to a number of important nonlinear systems.

\section{Forecasting Model}

We consider the use of a model driven by $K$ linear features
$$v_{k,t} = \sum_{\tau=1}^T \phi_{k,\tau} w_{t-\tau}, \qquad k = 1,\ldots, K.$$
Here, each $\phi_{k,\tau}$ is a scalar constant and each $v_{k,t}$ represents a scalar feature that is useful for predicting the next disturbance.  We consider a model of the form
$$w_t = \sum_{k=1}^K r_k v_{k,t} + z_t,$$
where $r \in \mathbb{R}^K$ is a vector of model coefficients and $z_t \sim {\cal N}(0, \sigma_z^2)$ is i.i.d Gaussian noise.  Note that the conditional expectation of disturbance $w_t$ at time $t$ is linear in past disturbances.

It is straightforward to generate forecasts $\hat{w}_{t,t}, \hat{w}_{t,t+1}, \ldots\hat{w}_{t,t+M-1}$ given the coefficients $r$, features $\phi_1,\ldots,\phi_K$, and
past disturbances $w_{t-1}, \ldots, w_{t-T}$.  In particular, let $\hat{w}_{t,\ell} = w_\ell$ for $\ell < t$, and for $\ell = t, \ldots, t+M-1$, recursively compute
$$\hat{v}_{k,t,\ell} = \sum_{\tau=1}^T \phi_{k,\tau} \hat{w}_{t,\ell-\tau}, \qquad k = 1,\ldots, K,$$
and
$$\hat{w}_{t,\ell} = \sum_{k=1}^K r_k \hat{v}_{k,t,\ell}.$$
In the event that the data is generated by our model, each forecast produced in this way is a conditional expectation: $\hat{w}_{t,t+\tau} = {\E}_t[w_{t+\tau}]$.

Since, conditioned on $w_{t-1}, \ldots, w_{t-T}$, the forecasts $\hat{w}_{t,t}, \ldots, \hat{w}_{t,t+M-1}$ are conditionally independent of previous disturbances, there is a function $F_r$ which maps the $T$ most recent disturbances to forecasts of the next $M$ disturbances.  In particular, 
$$\hat{w}_{t,t}^{t,t+M-1} = F_r(w_{t-T}^{t-1}),$$
where $w_{t-T}^{t-1} = [w_{t-T} \ \cdots \ w_{t-1}]^\top$. Note that $F_r$ depends on the coefficient vector $r$ and is generally not linear in $r$. From (\ref{eq:mu}) we see that
control actions generated by certainty equivalent MPC based on forecasts produced by our model can be written as
\begin{equation} \nonumber
u_t = L x_t + H F_r( w_{t-T}^{t-1} ).
\end{equation}
Let $\mu_r$ denote a control policy that maps $x_t$ and $w_{t-T}^{t-1}$ to $u_t$.  Note that this control policy depends on $r$ and again is generally not linear in $r$.

\section{Time Series Regression Algorithms}

We now discuss algorithms for computing coefficients to fit a forecasting model of the kind we have described to a time series. In particular, each algorithm will compute a vector  $r \in \mathbb{R}^K$ given observed samples $w_0, \ldots w_{N-1}$.

\subsection{Least Squares}

The most common approach is least squares regression (LS).  Here, we compute coefficients
\begin{equation} \nonumber
r^{\rm LS} :=  \argmin_r \sum_{t=T}^{N-1} \left\| w_t-\sum_{k=1}^K r_k v_{k,t} \right\|_2^2.
\end{equation}
Note that the sum begins with $t=T$ because earlier samples $w_0,\ldots,w_{T-1}$ are required to compute the features $v_{k,T}$. The optimization problem can be solved efficiently since it is a linear least squares problem. The resulting coefficient vector $r^{\rm LS}$ represents the maximum likelihood estimate assuming that the observed time series is generated by our forecasting model.  As such, this is a natural approach to fitting a forecasting model that accurately captures
the generating process.

\subsection{Empirical Optimization}

LS does not take the control problem into account when computing coefficients.  This assumes a separation principle whereby model fitting can be decoupled from the subsequent selection of control actions.  Empirical optimization (EO) provides an alternative that does account for the control problem when computing coefficients.  The algorithm computes coefficients
that minimize ``historical cost'':
\begin{equation}
r^{\rm EO} := \argmin_r \sum_{t=T}^{N-1} g(x^{\mu_r}_t, \mu_r(x^{\mu_r}_t,w_{t-T}^{t-1}) ), \label{eq:EO} 
\end{equation}
where $x^{\mu_r}_T$ is set to the initial state of the system  and subsequent control actions $u_T, u_{T+1}, \ldots, u_{N-1}$ are selected by $\mu_r$.  The objective here is the sum of costs that would have been realized if we were to apply certainty equivalent MPC alongside our forecast model with coefficients $r$, starting at time $T$. It is easy to see that if the disturbance process is ergodic, as the number $N$ of observed samples increases, historical average cost converges to future average cost, and therefore, EO computes coefficients that optimize future average cost.  However, for reasonable values of $N$, EO tends to overspecialize to the data, and this results in poor future performance.

It may appear difficult to compute $r^{\rm EO}$ because $\mu_r$ is generally not linear in $r$.  However, as in the case of the computational study we will later discuss, we have had positive 
experience applying local optimization methods initialized with $r^{\rm LS}$. A more detailed description of our implementation is given in the appendix.

Nevertheless, this non-convex optimization problem can still be challenging as the number of parameters $K$ increases, mainly because evaluating the gradients and Hessians of $F_r$ requires iterative computation. We now propose an approximation algorithm that relieve this problem by linearization. Note that $\mu_r$ is typically close to linear in a region of $\mathbb{R}^K$ that includes reasonable choices of $r$, for reasons we now explain.  The policy $\mu_r$ can be written as a sum of terms that are linear and terms that are nonlinear in $r$. Linear terms are multiples of terms of the form $r_k \phi_{k,\tau}$.  Nonlinear terms are higher order terms that involve products of terms of the form $r_k \phi_{k,\tau}$.  In most problems of practical interest, terms of the form $r_k \phi_{k,\tau}$ are significantly smaller than one because past disturbances exhibit decaying influence on future ones.  As such, the nonlinear terms tend to be significantly smaller than the linear ones, which therefore play a dominant role in $\mu_r$. This motivates the following approximation algorithm.

Suppose we have a reasonable estimate $\hat{r}$ for the model coefficients, which we will refer to as base coefficients. In practice, a typical choice for it is $r^{\rm LS}$. Given $\hat{r}$, it is straightforward to iteratively generate forecasts $\hat{w}_{t,t}, \hat{w}_{t,t+1}, \ldots, \hat{w}_{t,t+M-1}$ and features $\hat{v}_{k,t,t}, \hat{v}_{k,t,t+1}, \ldots, \hat{v}_{k,t,t+M-1}$, $k=1,\cdots, K$. Now fix these features and define
$$ \tilde{w}_{t,\ell} := \sum_{k=1}^K r_k \hat{v}_{k,t,\ell}, \quad \ell = t, \ldots, t+M-1.$$
In other words, here we fix base coefficients $\hat{r}$ when rolling forward but allow another set of coefficients $r$ to take effect at the last step of prediction. Apparently $\tilde{w}_{t,\ell}$ is linear in $r$. 
Furthermore, by similar notation we define
$$ \tilde{F}_{\hat{r},r} (w_{t-T}^{t-1}) := \tilde{w}_{t,t}^{t,t+M-1} $$
and
$$ \tilde{\mu}_{\hat{r}, r}(x_t,  w_{t-T}^{t-1}) := L x_t + H \tilde{F}_{\hat{r}, r} ( w_{t-T}^{t-1} ), $$
both of which are also (affine) linear in $r$. As we can see, this approximation trades in some model flexibility for linearity. This leads to an alternative formulation for EO, which we refer to as linearized EO (LEO):
\begin{equation}
r^{\rm LEO} := \argmin_r \sum_{t=T}^{N-1} g(x^{\tilde{\mu}_{\hat{r},r}}_t, \tilde{\mu}_{\hat{r},r} (x^{\tilde{\mu}_{\hat{r},r}}_t,w_{t-T}^{t-1}) ). \label{eq:LEO} 
\end{equation}
Unlike (\ref{eq:EO}), (\ref{eq:LEO}) can be solved efficiently by quadratic programming. A policy function $ \tilde{\mu}_{\hat{r}, r^{\rm LEO}}$ can then be built upon $r^{\rm LEO}$.

\subsection{Directed Time Series Regression}

Directed time series regression (DTSR) aims to combine the merits of LS and EO.  Directed regression was first introduced in \citet{Kao09} in a context involving repeated independent decision problems as opposed to intertemporal control. That work also developed a theory that motivated the algorithm and demonstrated its benefits through a computational study. What we consider here is an extension of directed regression that is suitable for time series.

A naive version of DTSR (NDR) produces a vector of coefficients that is a convex combination of those computed by LS and EO:
\begin{equation} \nonumber
	r^{\rm NDR} := (1-\lambda) r^{\rm LS} + \lambda r^{\rm EO},
\end{equation}
where $\lambda \in [0,1]$ is selected by cross-validation. This algorithm, albeit simple and intuitive, requires solving EO as an intermediate step and therefore can be inefficient for complicated problems. This disadvantage motivates the following approximation algorithm, linearized DTSR  (LDR), which produces a coefficient vector by taking a convex combination of $r^{\rm LS}$ and $r^{\rm LEO}$
\begin{equation} \nonumber
	r^{\rm LDR} := (1-\lambda) r^{\rm LS} + \lambda r^{\rm LEO}.
\end{equation}
A policy function $\tilde{\mu}_{\hat{r}, r^{\rm LDR}}$ then follows.

\begin{figure*}[ht]
\vskip 0.2in
\begin{center}
\centerline{\includegraphics[scale=0.4]{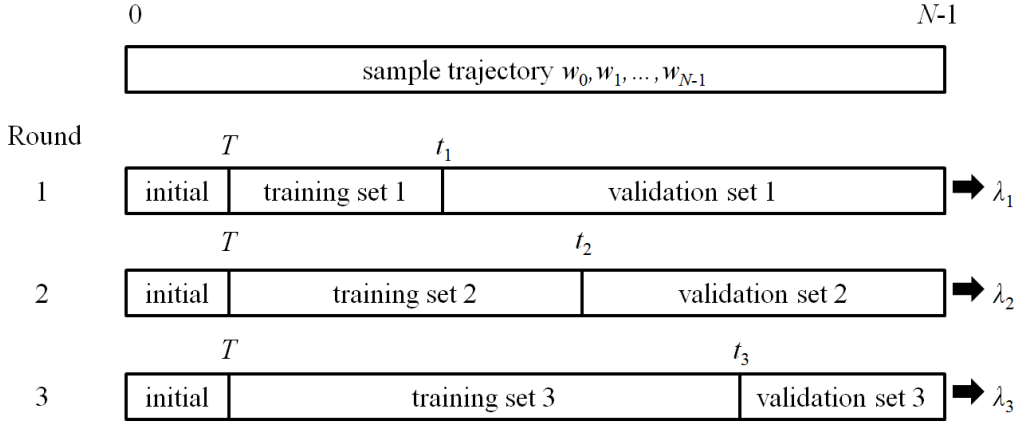}}
\caption{Sliding window cross-validation.}
\label{fig:cv}
\end{center}
\vskip -0.2in
\end{figure*}

We will use a sliding window cross-validation procedure to select the parameter $\lambda \in [0,1]$ for NDR and LDR. The windows are defined by a sequence $t_i \in \{T+1, T+2, \ldots, N-1\}$, each element specifying a boundary that separates the $i$th training set window from the $i$th validation set window.  To select the parameter $\lambda$ for NDR, for each $i$ we compute $r^{\rm LS}$ and $r^{\rm EO}$ using the observations $w_0, w_1, \ldots, w_{t_i}$, and then evaluate each choice of $\lambda_i$ by simulating the system with control actions generated by $\mu_{ (1-\lambda) r^{\rm LS} + \lambda r^{\rm EO}}$, starting at
state $x_{t_i} = 0$.  The performance of each $\lambda_i$ is judged based on the sum of costs incurred over time $t_i+1, \ldots, N-1$.  For each $i$, the best value of $\lambda_i$ is selected, and we take $\lambda$ to be the average over $i$ among the selected values of $\lambda_i$.  In our computational study, we set $t_1$, $t_2$, and $t_3$ to 
$T+0.3(N-T), T+0.5(N-T)$, and $T+0.7(N-T)$, each rounded off to the closest integer. The cross-validation procedure for LDR is essentially the same, except the replacement of $r^{\rm EO}$ by $r^{\rm LEO}$ and $\mu_{ (1-\lambda) r^{\rm LS} + \lambda r^{\rm EO}}$ by $\tilde{\mu}_{\hat{r}, (1-\lambda) r^{\rm LS} + \lambda r^{\rm LEO}}$. Figure \ref{fig:cv} illustrates this sliding window cross-validation procedure.

\section{Computational Study}

To assess relative merits of the algorithms we have discussed, we conducted a computational study involving an inverted pendulum problem of the kind described in Example 1.

\subsection{System Dynamics}

We discretize time, sampling at a rate of 100Hz. We employ a linearization of this system around the balance point, as derived in \citet{Landry05}. We also use the same physical parameters therein, and we assume that at each time $t$ there is a gust of wind accelerating the pendulum's angle by $w_t$.  Specifically, the system dynamics are characterized by the following matrices:
$$ A = \left[ \begin{array}{cccc}
1 & 0 & 0.01 & 0 \\
0 & 1 & 0  & 0.01 \\
0 & -0.0178 & 0.8872 & 0 \\
0 & 0.2847 & 0.2773 & 1 \\
\end{array} \right], $$
$$B = \left[ \begin{array}{c}
         0 \\
         0 \\
    0.0198 \\
   -0.04871 \\
\end{array} \right], \mbox{ }
C = \left[ \begin{array}{c}
	0 \\
	0 \\
	0 \\
	0.01 \\
\end{array} \right].
$$
Since our goal is to keep the cart centered and the pendulum vertical while minimizing energy consumption, we use a cost function parameterized by the following matrices:
$$ G_1 = \left[ \begin{array}{cccc}
1000 & 0 & 0 & 0 \\
0 & 1000 & 0  & 0 \\
0 & 0 & 1 & 0 \\
0 & 0 & 0 & 1 \\
\end{array} \right], G_2=0, G_3=0.1.$$
Hence, the cost in period $t$ is $x_t^\top G_1 x_t + 0.1 u_t^2$, where
the state is $x_t = [s_t \ \ \theta_t \ \ \dot{s}_t \ \ \dot{\theta}_t]^\top$ and the action $u_t$ influences acceleration. Along with the matrices $A, B, C, G_1, G_2$, and $G_3$, we take $M$ to be $100$ in our experiment and evaluate the policy matrices $L$ and $H$ accordingly.

\subsection{Generative and Forecasting Models}

In our study, we randomly sample an ensemble of generative models, each of which is used to generate a time series to which our algorithms are applied.  Each generative model is sampled as follows:
\begin{enumerate}
	\item Sample $\psi_1, \psi_2, \ldots, \psi_5$ i.i.d from ${\cal N}(0,1)$.
	\item Sample $\psi_6, \psi_7, \ldots, \psi_{30}$ i.i.d from ${\cal N}(0,\frac{1}{10^2})$.
	\item Select $\beta \in \mathbb{R}_+$ such that the process
				$$ w_{t} = \beta \sum_{\tau=1}^{30} \psi_\tau w_{t-\tau} + z_t $$
	has a variance $\E[w_t^2] = 2 \sigma_z^2$.
	\item Select $\sigma_z^2$ such that when control actions are selected according to $u_t = L x_t$, the average expected cost is equal to one.
\end{enumerate}
These steps result in an AR(30) process for which the first five coefficients tend to be an order of magnitude larger than the next twenty-five. The last two steps in the sampling process deserve some explanation.  Step 3 normalizes the signal-to-noise ratio of the disturbance process. This
reduces variations among sampled generative models in how helpful a forecasting model can be.  Step 4 serves to normalizes the control objective which would otherwise dramatically differ from one generative model to another. The control policy $u_t = L x_t$ is one that is optimal when future disturbances are not forecastable. The disturbance variance $\sigma_z^2$ is chosen so that the average expected cost when the system is controlled by this naive policy is equal to one.

Since the five most recent disturbance dominate others in influencing future disturbances, it is natural to select them as features of a forecasting model. In particular, we consider a forecasting model with five features:
$$v_{k,t} =  w_{t-k}, \qquad k = 1, \ldots, 5.$$
This model approximates the generative model but at the same time neglects useful information that can be extracted from disturbances observed beyond five time periods in the past.  It is our intention to consider a forecasting model that does not perfectly capture the generative model since it is in such contexts that DTSR adds value.  It is also important to recognize that this sort of model misspecification is inevitable in practical applications.

\subsection{Results}

For our computational study, we sampled 2,000 generative models using the above procedure.  For each model, a sequence of 5,360 disturbances $w'_0, \ldots, w'_{5359}$ are generated, initialized with an assumption that all disturbances prior to $w'_0$ are equal to zero.  We take the last $N$ samples of this trajectory as observations to be used by regression algorithms.  In other words, each algorithm makes use of disturbances $w_t = w'_{5360-N+t}$ for $t=0,1,\ldots, N-1$. In our experiment, we repeat the LS, EO, NDR, LEO, and LDR algorithms for each $N \in \{200, 240,\ldots, 360\}$.

To provide a lower bound, we consider a ``model-clairvoyant (MC) policy.''  This is the optimal policy to use given full knowledge of the generative model. The average expected cost $g^*$ incurred by this policy, which we will refer to as the MC cost, can be derived in closed form.  Clearly, policies generated by the aforementioned five algorithms must incur average expected costs no less than $g^*$.

\begin{figure}[ht]
\vskip 0.2in
\begin{center}
\centerline{\includegraphics[scale=0.9]{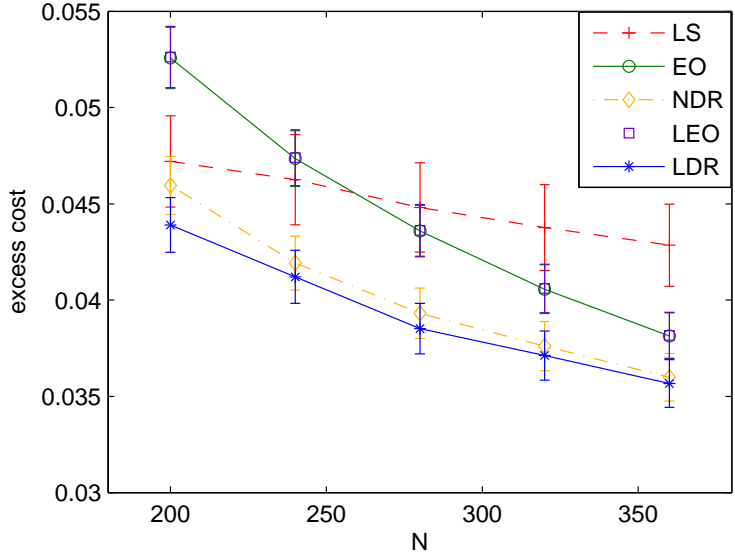}}
\caption{The excess costs delivered by LS, EO, NDR, LEO, and LDR for different numbers of training samples $N$; the curves of EO and LEO are essentially the same. Note that LDR is the dominant solution in all cases, reducing the excess costs delivered by LS/EO by as much as 12\%.}
\label{fig:loss}
\end{center}
\vskip -0.2in
\end{figure}

We can also evaluate in closed form the average expected costs $g^{\rm LS}$, $g^{\rm EO}$, $g^{\rm NDR}$, $g^{\rm LEO}$, and $g^{\rm LDR}$, of policies generated by our regression algorithms.  
We will focus our presentation on excess costs $g^{\rm LS} - g^*$, $g^{\rm EO} - g^*$, $g^{\rm NDR} - g^*$, $g^{\rm LEO} - g^*$, and $g^{\rm LDR} - g^*$. Subtracting $g^*$ factors out the influence of costs that are inevitable regardless of the choice of control policy. Figure \ref{fig:loss} plots excess costs as a function of $N$.  These results are qualitatively similar to those reported in \citet{Kao09} in a context involving repeated independent decision problems rather than intertemporal control. For small $N$, EO and LEO suffer from overspecialization.
For large $N$, the degree of overspecialization diminishes and the bias in LS due to model mispecification prevents LS from doing as well as EO and LEO. LDR always outperforms the best of these three methods, sometimes by as much as 12\%. It is also interesting to note that LDR provides a consistent gain over NDR, in spite of the substantial saving in computation.

\begin{figure}[ht]
\vskip 0.2in
\begin{center}
\centerline{\includegraphics[scale=0.6]{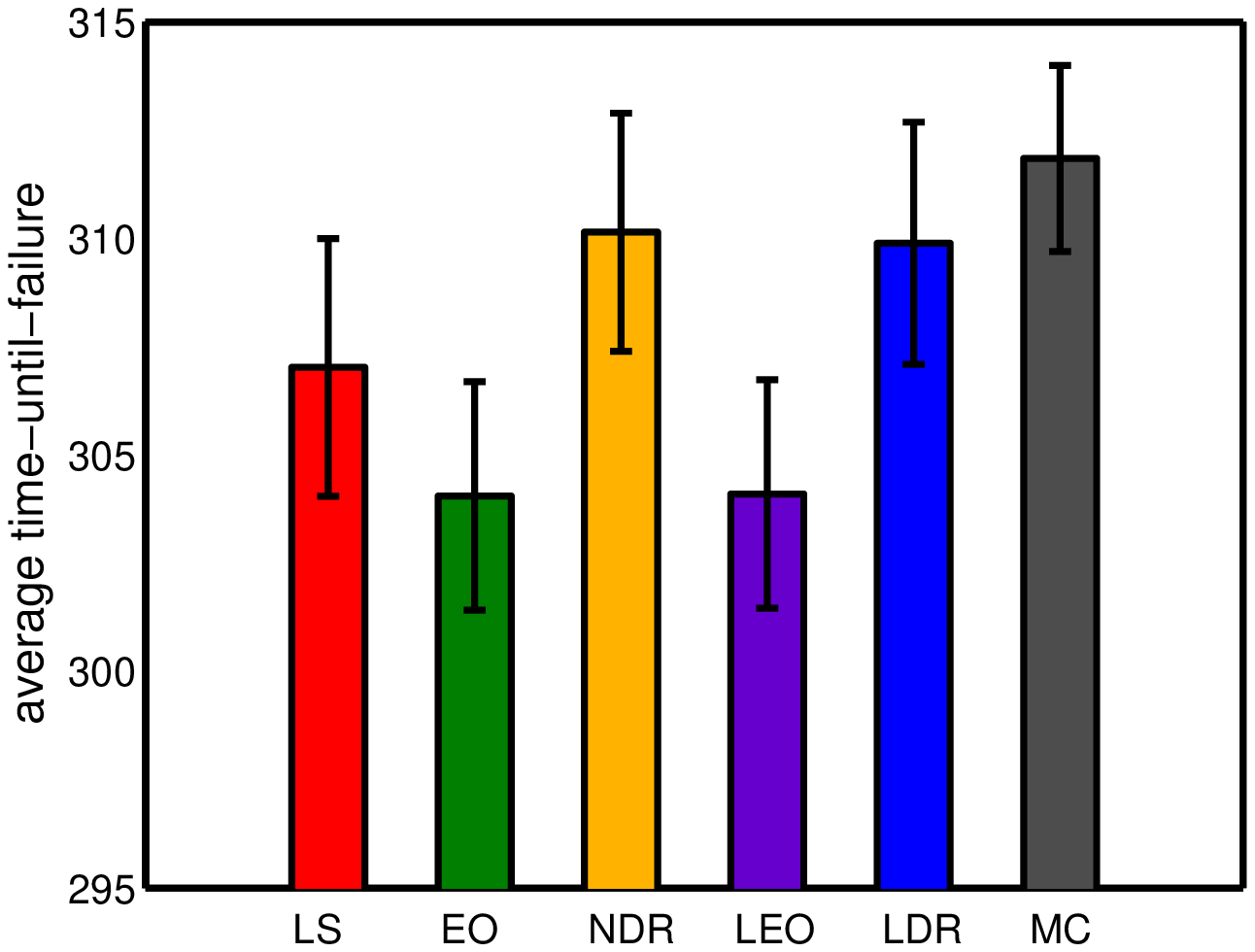}}
\caption{The average time-until-failure delivered by MC, LS, EO, NDR, LEO, and LDR, when a training data set of size 240 is given.  Note that NDR and LDR are much closer to MC than LS, EO, and LEO are.}
\label{fig:tuf}
\end{center}
\vskip -0.2in
\end{figure}

We can also compare quantities with simple physical interpretations. Suppose this cart-pendulum system is considered to reach a state of failure if the cart's position or the pendulum's angle deviates from zero by more than certain threshold values. Let the thresholds on position and angle be $B_s=0.0392$ and $B_\theta=0.0364$, which are roughly twice of the root-mean-square values of $s_t$ and $\theta_t$ in our simulations. We refer to the number of  time periods elapsed before a system initialized at the zero state reaches failure as \emph{time-until-failure}. For each of our 2,000 models, we generate 50 disturbance trajectories.  For each of these trajectories and each of the policies under consideration (MC, LS, EO, NDR, LEO, and LDR), we simulated the system until failure. Figure \ref{fig:tuf} summarizes results. It is clear that NDR and LDR are much closer to MC than LS, EO, and LEO are in terms of time-until-failure.

\section{Conclusion}

In this paper we have presented directed time-series regression, an algorithm that computes coefficients of a time-series model for use in certainty equivalent model predictive control. Two versions of this algorithm have been proposed, namely NDR and LDR, which we will collectively refer to as DTSR in the following discussion. We have shown that when the quantity of available data is limited and the forecasting model differs from the generative model, this algorithm offers a significant advantage by combining merits of least squares regression and empirical optimization. As side products, we have also proposed a methodology that transforms the original problem into a linearized version, and a sliding window cross-validation algorithm for time series with control. Both techniques can potentially enlarge the applicability of directed regression theory in other fields.

The main point of the paper is that the coordination of forecasting and control can be fruitful and we hope this observation will stimulate work along these lines involving broader classes of control problems and learning algorithms. Our approach extends to more complex settings than the one considered in this paper, for example, cases where disturbances are multidimensional and the system is nonlinear. It would also be interesting to explore the use of DTSR in contexts involving non-stationary time series.

Let us conclude with a discussion of several threads of work that relate to the ideas we have presented. Motivated by a similar perspective on fitting a mispecified model for use by a control algorithm, the approach developed in \citet{Abbeel05} learns a first-order Markov model in a way that improves controller performance when data is not generated by a first-order Markov model.
This approach takes the discount factor of the control problem into account when learning the transition matrix.  Aside from the contextual differences between discrete Markov processes and linear autoregressive models, a conceptual advance in our work can be seen in the fact that DTSR takes the entire control objective into account.

Differences between LS and EO are analogous to differences that have been studied between generative and discriminative methods for learning (see, e.g., \citet{Ng02}). When data samples are scarce, generative methods often provide better results, as does LS. On the other hand, when there is ample data, discriminative methods are advantageous, as is the case for EO. DTSR provides a useful way of combining the merits of LS and EO, and the idea may generalize to offer an approach that can more broadly be used to combine merits of generative and discriminative methods.

Control theorists working on system identification typically adopt weighted least-square linear regression to fit a linear system \citep{Ljung98}. While this approach puts more emphasis on learning the critical components, it does not explicitly consider the control objective. Econometricians have analyzed properties of parameter estimates for misspecified time series models \citep{White80,Domowitz82}.  However, this line of work does not treat the use of such models and estimated parameters for decision or control. Operations researchers have developed methods that factor objectives into how point estimates such as forecasts are generated when they are to be used for decision or control \citep{Granger69,Liyanage05,Chua08}.  However, this line of work does not consider model misspecification.

\section*{Appendix}

\subsection*{Implementation Details for EO}
The particular local optimization method we use in our computational study is a variation of the Guass-Newton method \citep{Bertsekas99}.  This suits our problem well because $g$ is quadratic.  The particular variation we use deviates from the standard Gauss-Newton method in that each iteration carries out a backtracking line search to determine the size of the step taken toward what would be the next Gauss-Newton iterate. Further, whenever the approximate Hessian is not positive definite, we add to it a small multiple of the identity matrix, in the spirit of the Levenberg-Marquardt method. In our computational study, the backtracking line search starts with a step size of one and halves the step size repeatedly so long as that improves the objective value in (\ref{eq:EO}).  When a multiple of the identity matrix is added, which is rare, the multiplier is $0.001$.

\bibliography{DTSR-ref-01-20-2010}

\end{document}